\documentclass[sigconf,nonacm]{acmart}
\AtBeginDocument{%
  }

\setcopyright{acmlicensed}
\copyrightyear{2026}
\acmYear{2026}
\acmDOI{XXXXXXX.XXXXXXX}
\acmConference[SIGKDD 2026]{Make sure to enter the correct
  conference title from your rights confirmation email}{August 9 to 13, 2026}{Jeju, Korea}
\acmISBN{978-1-4503-XXXX-X/2026/08}

\usepackage{algorithm}
\usepackage{algorithmicx,tikz}
\usepackage{algcompatible}
\usepackage{multirow}
\usepackage{float}

\begin{document}

\title{PowerGrow: Feasible Co-Growth of Structures and Dynamics for Power Grid Synthesis}

\author{Xinyu He}
\affiliation{%
  \institution{University of Illinois at Urbana-Champaign}
  \city{Champaign}
  \country{USA}}
\email{xhe34@illinois.edu}

\author{Chenhan Xiao}
\affiliation{%
  \institution{Arizona State University}
  \city{Tempe}
  \country{USA}}
\email{cxiao20@asu.edu}

\author{Haoran Li}
\affiliation{%
  \institution{Massachusetts Institute of Technology}
  \city{Cambridge}
  \country{USA}}
\email{haorandd@mit.edu}

\author{Ruizhong Qiu}
\affiliation{%
  \institution{University of Illinois at Urbana-Champaign}
  \city{Champaign}
  \country{USA}}
\email{rq5@illinois.edu}

\author{Zhe Xu}
\affiliation{%
  \institution{Meta}
  \city{Sunnyvale}
  \country{USA}}
\email{zhexu@meta.com}

\author{Yang Weng}
\affiliation{%
  \institution{Arizona State University}
  \city{Tempe}
  \country{USA}}
\email{yweng2@asu.edu}

\author{Jingrui He}
\affiliation{%
  \institution{University of Illinois at Urbana-Champaign}
  \city{Champaign}
  \country{USA}}
\email{jingrui@illinois.edu}

\author{Hanghang Tong}
\affiliation{%
  \institution{University of Illinois at Urbana-Champaign}
  \city{Champaign}
  \country{USA}}
\email{htong@illinois.edu}

\renewcommand{\shortauthors}{Xinyu et al.}

\begin{abstract}
Modern power systems are becoming increasingly dynamic, with changing topologies and time-varying loads driven by renewable energy variability, electric vehicle adoption, and active grid reconfiguration.
Despite these changes, publicly available test cases remain scarce, due to security concerns and the significant effort required to anonymize real systems. Such limitations call for generative tools that can jointly synthesize grid structure and nodal dynamics. 
However, modeling the joint distribution of network topology, branch attributes, bus properties, and dynamic load profiles remains a major challenge, while preserving physical feasibility and avoiding prohibitive computational costs.
We present PowerGrow, a co-generative framework that significantly reduces computational overhead while maintaining operational validity. The core idea is dependence decomposition: the complex joint distribution is factorized into a chain of conditional distributions over feasible grid topologies, time-series bus loads, and other system attributes, leveraging their mutual dependencies. By constraining the generation process at each stage, we implement a hierarchical graph beta-diffusion process for structural synthesis, paired with a temporal autoencoder that embeds time-series data into a compact latent space, improving both training stability and sample fidelity.
Experiments across benchmark settings show that PowerGrow not only outperforms prior diffusion models in fidelity and diversity but also achieves a 98.9\% power flow convergence rate and improved N-1 contingency resilience. This demonstrates its ability to generate operationally valid and realistic power grid scenarios. 

\end{abstract}





\maketitle

\section{Introduction}

Modern power grids are undergoing rapid structural and temporal changes due to the integration of renewable energy, electric vehicles, and demand-side technologies~\cite{fischer2020impact, wang2018review}. These trends introduce significant variability in both network topology and nodal power injections, requiring accurate models to simulate system-wide behavior under evolving conditions. However, publicly available test cases remain scarce due to data sensitivity and the substantial effort needed to anonymize real systems and construct realistic scenarios \cite{zimmerman2010matpower, bu2019time}. As a result, synthetic datasets are increasingly used to support forecasting, planning, and resilience studies. 

There is extensive existing work on producing synthetic grid structures and time-series dynamics, such as load profiles~\cite{fernandez2022privacy, wang2022generating}.. For example, traditional load and topology synthesis methods often rely on statistical models~\cite{birchfield2016grid,birchfield2016statistical,proedrou2021comprehensive}, which fail to capture the complex spatial-temporal dependencies in modern power systems. While deep learning models such as Graph Neural Networks (GNNs) \cite{wu2020comprehensive} and Long Short-Term Memory (LSTM) networks \cite{graves2012long} have demonstrated promise in learning from structured and sequential data, they typically disregard physical laws and operational feasibility constraints. As a result, outputs from these models often lack fidelity or violate grid constraints, limiting their applicability in real-world settings~\cite{proedrou2021comprehensive}. In general, existing approaches usually treat topology generation and load profile synthesis as two separate tasks~\cite{yan2022active, shahraeini2023modified, khodayar2019deep, song2025synthetic}.

However, time-series operational data and graph structures have strong interdependence due to inherent topology consistency \cite{dande2024synthetic, giacomarra2024generating, hu2023multiload, pinceti2021synthetic}. For example, in operational practice, power systems are continuously reconfigured both topologically and temporally to ensure reliability and efficiency ~\cite{tang2017real}. Hence, it's important to align and evaluate these two types of generations together. Sequential pipelines that first generate a grid and then simulate loads have been proposed~\cite{pinceti2019data}, but they fail to enforce physical consistency: generated loads may violate feasibility constraints under the assumed topology, and the topology itself may not reflect the influence of underlying demand patterns~\cite{molzahn2017survey}. This fact leads to infeasible operating points, voltage violations, and artificial use cases that are highly unlikely to happen, which compromise planning, simulation, or control tasks.

Therefore, some cutting-edge work \cite{yan2021synthetic} introduces post-process efforts, e.g., conducting mixed integer programming, to correctly combine generated graph structures, physical parameters, and load curves into a feasible and operable system. This optimization-based procedure is computationally expensive, which can not scale to large systems. 
In this paper, we explore the strong correlations between graphs and time series to explore conditional dependence to reduce the generation space for reduced time.  

Specifically, we investigate diffusion models for power network because they offer a compelling foundation: recent advances in graph diffusion models~\cite{GDSS, GruM, EDP-GNN, DISCO} have demonstrated strong performance across various structured domains such as molecule design \cite{hua2024mudiff, liu2024graphxxx}, social networks, etc. However, as shown in our resulting Table ~\ref{tab:exp:mmd}, these methods meet significant performance degradation when directly applied to complicated physical systems, such as power systems. 
First, unlike domains such as molecular generation where node and edge features are typically discrete (e.g., atom and bond types) \cite{hua2024mudiff}, power systems exhibit a mix of tightly coupled discrete (e.g., bus types, topology connections), continuous (e.g., line impedance, power injections), and temporal data (e.g., nodal load profiles). Modeling such heterogeneous data in a unified way poses a high-dimensional, interdependent generation problem that overwhelms standard end-to-end diffusion architectures. Second, nodal load profiles in power systems span long temporal horizons and exhibit complex dependencies, including daily and weekly cycles~\cite{hodge2020hourly}, spatial correlations among neighboring buses~\cite{li2023power}, etc. A generative model should also accurately capture it. Third, there is a feasibility concern for the above multi-type data generations. 

To manage this complexity, we propose a hierarchical Graph Beta Diffusion framework that employs a conditional generation strategy: discrete components such as bus types and structural connectivity are synthesized first, followed by continuous attributes like branch impedances and temporal attributes like load profiles. This coarse-to-fine factorization aligns with real-world grid formation and significantly improves generation fidelity and feasibility. Moreover, to efficiently capture temporal features, we pretrain a Long Short-Term Memory Autoencoder (LSTM-AE) to extract compact latent representations. These embeddings serve as dynamic node features that are integrated into the diffusion-based structural generation, enabling coherent modeling across both spatial and temporal dimensions. Then, the diffusion model and the decoder in LSTM-AE are fine-tuned to extract the interdependence and achieve the alignment between the graph and the time series. Finally, to achieve high feasibility, we employ a domain-specific power flow simulator \cite{wood2013power,emanuel2011power} to prepare high-quality training data.

Overall, a three-level hierarchical graph beta diffusion framework is proposed to decompose the joint generation task into three conditional generation subtasks. We term this three-level diffusion framework as \textbf{PowerGrow}, whose pipeline is shown in Fig. \ref{fig:bigpic}. In summary, our main contributions are as follows:

\begin{itemize}
\item \textbf{Joint Structural–Temporal Synthesis}: We tackle the underexplored task of co-generating realistic grid topologies and nodal load profiles, explicitly modeling their mutual constraints to ensure operational feasibility without costly post-processing.

\item \textbf{Hierarchical Dependence Decomposition}: We propose PowerGrow, a three‑stage Graph Beta Diffusion framework that factorizes generation into structure, branch attributes, and temporal loads, mirroring real‑world grid formation and improving scalability and fidelity.

\item \textbf{Efficient Long‑Horizon Load Modeling}: We integrate an LSTM‑based temporal autoencoder to compress load profiles into compact node embeddings, reducing generation cost and preserving temporal coherence.

\item \textbf{Strong Empirical Results}: On benchmark systems, PowerGrow achieves <0.01 MMD scores, 98.9\% power flow convergence, and improved N‑1 resilience, producing operationally valid grids that outperform prior generative models in both fidelity and diversity.
\end{itemize}

\vspace{-1em}
\section{Related Work}

{\bf Power grid topology and load synthesis.} 
Generating synthetic power grid topologies and load profiles is becoming increasingly important for scalable analysis, system design, and planning \cite{yan2022active, shahraeini2023modified, khodayar2019deep, song2025synthetic, dande2024synthetic, giacomarra2024generating}. Early topology synthesis used VAEs to model grid structure \cite{khodayar2019deep}, later methods extended to preserve structural motifs \cite{giacomarra2024generating} and incorporate geographic constraints \cite{song2025synthetic, dande2024synthetic}. In parallel, load generation has leveraged GANs to capture spatio-temporal patterns in historical data \cite{hu2023multiload, pinceti2021synthetic, yan2022active}.

However, most existing approaches treat topology and load synthesis as decoupled tasks \cite{yan2022active, shahraeini2023modified, khodayar2019deep, song2025synthetic}, ignoring their physical interplay \cite{dande2024synthetic, giacomarra2024generating, hu2023multiload, pinceti2021synthetic}. This can lead to infeasible grids \cite{molzahn2017survey}. For example, generating a weakly connected topology followed by heavy industrial loads may violate voltage or thermal limits, while overlaying low-variability residential loads onto a high-redundancy mesh grid leads to overdesign and inefficiency. We bridge this gap by proposing a hierarchical co-generation framework that integrates multi-level Graph Beta Diffusion (GBD) with a temporal autoencoder. This joint modeling explicitly captures the structural-temporal interplay.

{\noindent \bf Graph diffusion generative models.} Graph diffusion generative models have attracted growing interest recently owing to their strong sample fidelity and stable training. Early efforts such as EDP-GNN~\cite{EDP-GNN} and GDSS~\cite{GDSS} model graph data in a continuous state space, requiring quantization during sampling to recover discrete structures. To better align with the inherently discrete nature of graphs, methods like DiGress~\cite{DiGress} and DisCo~\cite{DISCO} adopt discrete-state diffusion processes, where edges and node types evolve categorically during noise corruption. Several extensions explore more flexible diffusion paradigms: EDGE~\cite{EDGE} diffuses the clean graphs to empty graphs, different from typical random graphs; GraphArm~\cite{GraphArm} introduces absorbing states through masked nodes and edges, framing the diffusion process in an autoregressive manner; Graph Beta Diffusion~\cite{GBD} applies a beta diffusion process that accommodates both discrete graph structures and continuous node attributes.

Although graph diffusion models are widely used in fields such as bioinformatics~\cite{guo2024diffusion,huang2023conditional,hoogeboom2022equivariant,waibel2023diffusion} and material science~\cite{yangscalable,xiecrystal,kelviniuswyckoffdiff}, applying them directly to power grid generation is nontrivial. Power grid data combines discrete topology with continuous node/edge attributes and temporal features. Generating synthetic grid graphs that maintain physical feasibility, discrete structure, and temporal coherence remains a complex and largely unexplored problem.




\vspace{-1em}
\section{Preliminaries}


{\bf Notations and problem formulation.} 
We represent a power grid as a graph $\mathcal{G} = (X, A, E, D)$, where:
$X\in \mathbb{R}^{N \times d_1}$ denotes the $d_1$-dimensional node features for $N$ nodes (buses), including the node types (generator bus or load bus) and nodal constraint parameters, such as maximum generation limits, ramp rates, and unit costs.
$A \in \mathbb{R}^{N \times N}$ is the adjacency matrix specifying the grid topology. In alignment of the shape of the adjacency matrix,
we formulate edge (branch) features as $E\in \mathbb{R}^{N\times N\times d_2}$, where $E[i,j]$ denotes the $d_2$-dimensional feature of the branch between bus $i$ and $j$, including electrical parameters (e.g., resistance, reactance) and operational constraints (e.g., line power flow limits).
Finally, we denote by $D \in \mathbb{R}^{N \times T\times d_3}$ the nodal-level time-series matrix containing power grid load profile over $T$ time steps.

This framework is flexible and can be extended to incorporate additional node- or branch-level features. For instance, $X$ may further include voltage setpoints, while $E$ can be augmented with parameters for advanced transmission components such as series compensation devices or flexible AC transmission systems (FACTS), which are commonly used to mitigate resonance and improve power flow controllability.

Let $\mathbf{v}$ denote the vector of physical and operational constraint violations defined over buses and branches  (detailed constraints discussed in Appendix~\ref{app:v:constraint}), our objective is to develop a generative model to generate $\tilde{\mathcal{G}}$ satisfying two key criteria: $(1)$ \emph{fidelity}: the distribution of $\tilde{\mathcal{G}}$ is close to that of real grids $\mathcal{G}$ and $(2)$ \emph{feasibility}: the topology and loads in $\tilde{\mathcal{G}}$ should have minimal violations in $\mathbf{v}$. In Section~\ref{sec:exp}, we further quantify the feasibility by mapping $\mathbf{v}$ to a continuous feasibility score that reflects the severity of constraint violations.

{\noindent \bf Graph beta diffusion.} Denoising diffusion models are composed of two processes: a forward diffusion process and a backward denoising process. In the forward process, the input data is gradually diffused into near-zeros or pure noise by applying noises sampled from a fixed noise distribution (e.g., gaussian distribution, beta distribution) step-by-step. Conversely, the reverse process learns to reconstruct the input data from noises using trainable neural networks.
Compared to Gaussian noise, Beta-distributed noise better models the characteristics of graph data relevant to power systems: it naturally captures structural sparsity, long-tailed feature distributions, and bounded attribute ranges \cite{GBD}. In graph beta diffusion \cite{GBD}, the forward multiplicative $K$-step beta diffusion process \cite{zhou2023beta} for a plain graph is formulated as: $\forall k\in \{1,2,\dots,K\}$,
\begin{equation}\label{eq:beta:forward:diffusion}
\begin{split}
    A_k &= A_{k-1}\odot Q_{A,k},\quad X_k = X_{k-1}\odot Q_{X,k},\\
    Q_{A,k} &\sim \mathtt{Beta}(\eta_A\alpha_kA_0, \eta_A(\alpha_{k-1}-\alpha_k)A_0), \\
    Q_{X,k} &\sim \mathtt{Beta}(\eta_X\alpha_kX_0, \eta_X(\alpha_{k-1}-\alpha_k)X_0),
\end{split}
\end{equation}
where $\mathtt{Beta}$ is beta distribution, $\eta_A$ and $\eta_X$ adjust the concentration of beta distributions, and $\alpha_k$ is the noise schedule at step $k$, $(A_0, X_0)=(A,X)$ are the input adjacency and feature matrices.
Based on the beta diffusion process defined in Eq.~\eqref{eq:beta:forward:diffusion}, the marginal distribution for forward process is 
\begin{equation}
\begin{split}
    q(A_k|A_0) = \mathtt{Beta}\left(\eta_A\alpha_kA_0, \eta_A(1-\alpha_kA_0)\right), \\
    q(X_k|X_0) = \mathtt{Beta}\left(\eta_X\alpha_kX_0, \eta_X(1-\alpha_kX_0)\right).
\end{split}
\end{equation}
To equivalently construct the trajectory in the reverse order, the reverse multiplicative sampling process is formulated as
\begin{equation}\label{eq:sample:add}
\begin{split}
    A_{k-1} &= A_k+\delta A_k,\quad X_{k-1} = X_k+\delta X_k\\
    \delta A_k &= P_{A,k}\odot (1-A_k),\quad \delta X_k = P_{X,k}\odot (1-X_k),\\
    P_{A,k}&\sim \mathtt{Beta}(\eta_A(\alpha_{k-1}-\alpha_k)A_0, \eta_A(1-\alpha_{k-1}A_0)), \\
     P_{X,k}&\sim \mathtt{Beta}(\eta_X(\alpha_{k-1}-\alpha_k)X_0, \eta_X(1-\alpha_{k-1}X_0)).
\end{split}
\end{equation}

The reverse process, defined via ancestral sampling, is:
\begin{equation}\label{eq:gbd:reverse}
\begin{split}
    p_\alpha(A_{k-1}|A_k) &= \frac{1}{1-A_k}\mathtt{Beta}\bigg(\frac{A_{k-1}-A_k}{1-A_k}\bigg|\eta_A(\alpha_{k-1}-\alpha_k)\hat A_0
    \\&\qquad\qquad\qquad\qquad, \eta_A(1-\alpha_{k-1}\hat A_0)\bigg), \\
    p_\alpha(X_{k-1}|X_k) &= \frac{1}{1-X_k}\mathtt{Beta}\bigg(\frac{X_{k-1}-X_k}{1-X_k}\bigg|\eta_X(\alpha_{k-1}-\alpha_k)\hat X_0
    \\&\qquad\qquad\qquad\qquad, \eta_X(1-\alpha_{k-1}\hat X_0)\bigg), \\
    \left(\hat A_0, \hat X_0\right) & = \mathbb{E}[A_0, X_0\mid A_k, X_k, k],
\end{split}
\end{equation}
where $\frac{1}{1-A_k}$ stands for element-wise inversion of $1-A_k$. In practice, a neural network $f_\alpha()$ is trained to predict the expectation of $\mathcal G_0$ given $\mathcal G_k$, $\mathbb{E}[A_0, X_0\mid A_k, X_k, k]\approx f_\alpha(A_k, X_k, k)$.


\begin{figure*}
    \centering
    \includegraphics[width=0.73\linewidth]{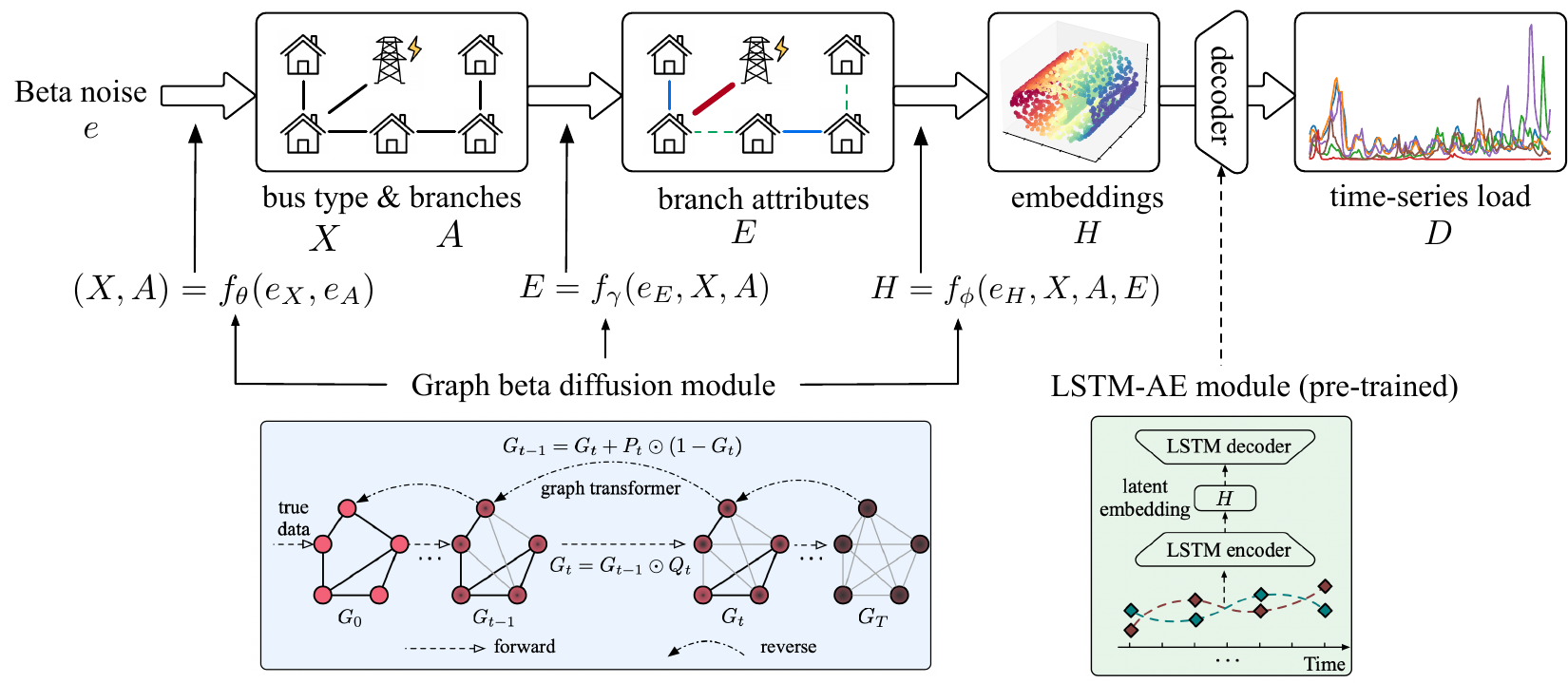}
    \vskip -0.1in
    \caption{Visualization of the sampling process of the proposed PowerGrow framework. The framework is composed of: (1) \textit{LSTM Autoencoder} module, and (2) \textit{three graph transformers} (graph beta diffusion module). The training of the three graph transformers is done in parallel. The encoder in pretrained LSTM-AE is frozen for encoding time-series, but the decoder is finetuned along with the training of diffusion module. For Sampling, $A$ and $X$, $E$, $H$, are sampled sequentially.}
    \label{fig:bigpic}
    \vskip -0.1in
\end{figure*}

\section{Method}
In this section, we first tackle the efficiency problem for lone-span time sequence diffusion through time series modeling in Section~\ref{sec:method:lstm}. Section~\ref{sec:method:beta} then introduces the hierarchical beta diffusion model for decomposing complex joint generation problem into three simpler conditional generation subtasks. Section~\ref{sec:method:framework} summarizes the training paradigm and overall framework of our PowerGrow algorithm.
\subsection{Time Series Modeling}\label{sec:method:lstm}
Loads in power grids often exhibit clear periodic patterns which can be captured by sequential models. In this work, PowerGrow enables a co-generation of bus loads and power grids topology, which involves the $4$ factorized distributions in the training of diffusion models. A direct adaptation of graph beta diffusion \cite{GBD} to joint power grid synthesis in forward process is formulated as
\begin{equation}\label{eq:gbd:forward}
\begin{split}
    q(A_k|A_0) &= \mathtt{Beta}\left(\eta_A\alpha_kA_0, \eta_A(1-\alpha_kA_0)\right), \\
    q(E_k|E_0) &= \mathtt{Beta}\left(\eta_A\alpha_kE_0, \eta_A(1-\alpha_kE_0)\right), \\
    q(X_k|X_0) &= \mathtt{Beta}\left(\eta_X\alpha_kX_0, \eta_X(1-\alpha_kX_0)\right). \\
    q(D_k|D_0) &= \mathtt{Beta}\left(\eta_X\alpha_kD_0, \eta_X(1-\alpha_kD_0)\right).
\end{split}
\end{equation}
where we assume that $A, E$ and $X, D$ share the same concentration $\eta_A$ and $\eta_X$, e.g., $A$ and $E$ should share the same sparsity. This forward diffusion process becomes increasingly expensive with the growth of $T$ from tens to thousands, because the complexity of diffusing $\{D_k\}_{k=1}^T$ is $O(NTd_3)$. To facilitate efficient diffusion model learning and generation, we are inspired by the latent diffusion~\cite{rombach2022high} to encode the power loads into a temporal embedding vector $H\in\mathbb{R}^{N\times d_4}$ via a sequential model, where $d_4=32$ in our experiments. Specifically, an LSTM-based autoencoder is applied:
\begin{equation}
    H = g_{\mathtt{enc}}(D),\quad \hat{D} = g_{\mathtt{dec}}(H),
\end{equation}
where $g_{\mathtt{enc}}$ and $g_{\mathtt{dec}}$ are two LSTMs.
We pretrain this LSTM-AE via reconstruction loss for two motivations: (1) to avoid the cold-start problem and ensure stable training: in early stage of training, $H$ does not hold meaningful information from $D$, which hinders the learning of diffusion model, and (2) avoid constructing extra large computational graphs from automatic differentiation engine (e.g., PyTorch) whose backpropagation is expensive. Therefore a three-stage pipeline is applied for training. First, LSTM-AE is trained in an unsupervised way with Mean Squared Error (MSE). Then diffusion model will take embeddings of dynamic features as inputs for generation, i.e., instead of diffusing $q(D_k|D_0)$, we will diffuse $q(H_k|H_0) = \mathtt{Beta}\left(\eta_X\alpha_kH_0, \eta_X(1-\alpha_kH_0)\right).$
Lastly, diffusion model and LSTM-AE will be finetuned alternatively to better align embedding space of LSTM-AE with hidden space of diffusion model. In this way, we ensure a more stable training process, and leverage the interdependency between bus loads and other factors.


\subsection{Hierarchical Beta Diffusion}\label{sec:method:beta}
Through ancestral sampling, a direct adaptation of reverse diffusion process involves a trainable network $f_\alpha$ for predicting
\begin{equation}
    \left(\hat{A}_0, \hat{X}_0, \hat{E}_0, \hat{H}_0\right) = f_\alpha(A_k, X_k, E_k, H_k, k)
\end{equation}
at diffusion step $k$. Previous graph diffusion models mainly focus on categorical features in molecule generation, or node-attributed graph. In this work, the co-generation of numerical edge features and time-series bus loads poses a new challenge for graph diffusion models. Our experiments (Table.~\ref{tab:exp:mmd}) show that existing methods fail to properly learn the dependencies between the four factors. To tackle this challenge, we decompose the complex joint generation task into three conditional generation tasks, which aligns with the hierarchical causal dependency between power structure and bus types, edge features, and bus loads.
We decompose the distribution $P_\alpha$ that $f_\alpha$ is trying to learn as
\begin{align}
    &\quad P_\alpha\left(\hat{A}_0, \hat{X}_0, \hat{E}_0, \hat{H}_0\big|A_k, X_k, E_k, H_k, k\right) \nonumber \\
    &= P_\phi\left(\hat{H}_0\big| \hat{A}_0, \hat{E}_0, \hat{X}_0, A_k, X_k, E_k, H_k, k\right) \cdot \nonumber\\
    &\quad P_\gamma\left(\hat{E}_0\big|\hat{A}_0, \hat{X}_0, A_k, X_k, E_k, H_k, k\right)\cdot P_\theta\left(\hat{A}_0, \hat{X}_0\big|A_k, X_k, E_k, H_k, k\right)\nonumber\\
    &= P_\phi\left(\hat{H}_0\big| \hat{A}_0, \hat{E}_0, \hat{X}_0, H_k, k\right) \cdot P_\gamma\left(\hat{E}_0\big|\hat{A}_0, \hat{X}_0, E_k, H_k, k\right) \cdot \nonumber\\
    &\quad P_\theta\left(\hat{A}_0, \hat{X}_0\big|A_k, X_k, E_k, H_k, k\right). \label{eq:p:alpha}
\end{align}

Here We assume that $A_k$ and $X_k$ contain no additional predictive information beyond $\hat{A}_0$ and $\hat{X}_0$ for estimating $\hat{E}_0$ and $\hat{H}_0$, as they represent noisy perturbations. A similar assumption applies to $E_k$ for predicting $\hat{H}_0$. Furthermore, we assume that the generation of $A, X$ is independent of $E, H$, and that $E$ is independent of $H$, consistent with the underlying hierarchical causal structure: topology and node types are first sampled, followed by branch attributes, and finally load profiles, each conditioned only on prior stages. Based on the independency assumption, Eq.~\eqref{eq:p:alpha} is simplified to 
\vspace{-0.5em}
\begin{align}
    P_\alpha\Big(\hat{A}_0, &\hat{X}_0, \hat{E}_0, \hat{H}_0\big|A_k, X_k, E_k, H_k, k\Big)
    = P_\phi\Big(\hat{H}_0\big| \hat{A}_0, \hat{E}_0, \hat{X}_0, H_k, k\Big) \cdot \nonumber\\
    &P_\gamma\Big(\hat{E}_0\big|\hat{A}_0, \hat{X}_0, E_k, k\Big)\cdot P_\theta\Big(\hat{A}_0, \hat{X}_0\big|A_k, X_k, k\Big). \label{eq:p:subtasks}
\vspace{-0.5em}
\end{align}
From Eq.~\eqref{eq:p:subtasks}, the joint generation task is naturally decomposed into three conditional generation subtasks: the generation of grid structure, generation of branch features conditioned on grid structure, and generation of bus loads conditioned on grid structure and branch features. Therefore, the trainable module in reverse diffusion process is formulated as
\vspace{-0.5em}
\begin{equation}
    \begin{split}
        \left(\hat{A}_0, \hat{X}_0\right) &= f_\theta(A_k, X_k,k), \\
        \hat{E}_0 &= f_\gamma(A_0, X_0, E_k, k), \\
        \hat{H}_0 &= f_\phi(A_0, X_0, E_0, H_k, k),
    \end{split}
\end{equation}
where $f_\theta, f_\gamma, f_\phi$ are implemented as graph transformers \cite{DiGress}.
This three-level decomposition allows us to train the three graph transformers independently and in parallel, reaching high training efficiency and stability. This also enables the application of PowerGrow in individual tasks, e.g., power grid synthesis given bus loads, or bus load generation given power grid. With this decomposition, we only need to finetune $f_\phi$ with LSTM-AE alternatively in the third stage of PowerGrow. 

Based on the three-level graph diffusion model, the reverse diffusion process is formulated as 
\begin{align}
    p_\theta(A_{k-1}|A_k) &= \frac{1}{1-A_k}\mathtt{Beta}\bigg(\frac{A_{k-1}-A_k}{1-A_k}\bigg|\eta_A(\alpha_{k-1}-\alpha_k)\hat A_0\nonumber
    \\&\qquad\qquad\qquad\qquad, \eta_A(1-\alpha_{k-1}\hat A_0)\bigg) \nonumber\\
    p_\theta(X_{k-1}|X_k) &= \frac{1}{1-X_k}\mathtt{Beta}\bigg(\frac{X_{k-1}-X_k}{1-X_k}\bigg|\eta_X(\alpha_{k-1}-\alpha_k)\hat X_0\nonumber
    \\&\qquad\qquad\qquad\qquad, \eta_X(1-\alpha_{k-1}\hat X_0)\bigg) \nonumber\\
    p_\gamma(E_{k-1}|E_k) &= \frac{1}{1-E_k}\mathtt{Beta}\bigg(\frac{E_{k-1}-E_k}{1-E_k}\bigg|\eta_A(\alpha_{k-1}-\alpha_k)\hat E_0
    \\&\qquad\qquad\qquad\qquad, \eta_A(1-\alpha_{k-1}\hat E_0)\bigg)\nonumber \\
    p_\phi(H_{k-1}|H_k) &= \frac{1}{1-H_k}\mathtt{Beta}\bigg(\frac{H_{k-1}-H_k}{1-H_k}\bigg|\eta_X\left(\alpha_{k-1}-\alpha_k\right)\hat H_0\nonumber\\
    &\qquad\qquad\qquad\qquad, \eta_X\left(1-\alpha_{k-1}\hat H_0\right)\bigg) \nonumber\\
    (\hat A_0, \hat X_0) &= f_\theta(A_k, X_k, k),\  \hat E_0 = f_\gamma\left(\hat A_0, \hat X_0, E_k, k\right),\nonumber\\ \hat H_0 &= f_\phi\left(\hat A_0, \hat X_0, \hat E_0, H_k, k\right).\nonumber
\end{align}
Note that $\hat A_0, \hat X_0, \hat E_0$ are treated as estimation of ground truth $A_0, X_0, E_0$ as those ground truth are not available in generation process.

\subsection{PowerGrow Framework}\label{sec:method:framework}
We employ the KLUB loss \cite{GBD, zhou2023beta} for training graph beta diffusion module. Take $X$ as an example, KLUB loss is consists of two parts, time reversal loss $\mathcal{L}_\text{sampling}$ between $p_\theta(X_{t-1}|X_t)$ and $q(X_{t-1}|X_t, X_0)$, and error accumulation control loss $\mathcal{L}_\text{correction}$ between $q(X_t|\hat{X}_0)$ and $q(X_t| X_0)$.
\begin{align}
    &\mathcal{L}_\text{sampling}(X_0, t) = \mathbb{E}_{q(X_t, X_0)}\text{KL}\left(p_\theta(X_{t-1}|X_t)\|q(X_{t-1}|X_t, X_0)\right) \nonumber\\
    &\mathcal{L}_\text{correction}(X_0, t) = \mathbb{E}_{q(X_t, X_0)}\text{KL}\left(q(X_t|\hat{X}_0)\big\|q(X_t| X_0)\right)  \nonumber\\
    &\mathcal{L} = \sum_{t=2}^T(1-\omega)\mathcal{L}_\text{sampling}(X_0, t) + \omega\mathcal{L}_\text{correction}(X_0, t). \label{eq:klub}
\end{align}

The time traversal loss optimizes the trainable graph transformers to match the ancestral sampling distribution with the ground truth conditional posterior distribution; the error accumulation control loss corrects the bias on the marginal distribution accumulated through the time steps. 
\cite{zhou2023beta} derives the KL-divergence between beta distributions as Bregman divergence between log-beta distributions. Specifically, the loss terms can be expressed as
\begin{align}
    &\mathcal{L}_\text{sampling}(X_0, t) = \mathbb{E}_{q(X_t, X_0)}[D_{\ln{\mathtt{Beta}(a,b)}}\{[\eta_X(\alpha_{t-1}-\alpha_t)X_0 \nonumber\\ 
    &\qquad\qquad, \eta_X(1-\alpha_{t-1}X_0)], [\eta_X(\alpha_{t-1}-\alpha_t)\hat X_0, \eta_X(1-\alpha_{t-1}\hat X_0)]\}] \nonumber \\
    &\mathcal{L}_\text{correction}(X_0, t) = \mathbb{E}_{q(X_t, X_0)}[D_{\ln{\mathtt{Beta}(a,b)}}\{[\eta_X\alpha_tX_0, \eta_X(1-\alpha_tX_0)]\nonumber\\ 
    &\qquad\qquad, [\eta_X\alpha_t\hat X_0, \eta_X(1-\alpha_t\hat X_0)]\}] 
\end{align}
where $D_f\{:,:\}$ is the Bregman divergence.

The overall training and sampling pipeline are summarized in Alg.~\ref{alg:train} and Alg.~\ref{alg:sample} in appendix. The pipeline is also visualized in Fig.\ref{fig:bigpic}.
In detail, we first train the LSTM Autoencoder module, then we train the three graph transformers (graph beta diffusion module), which could be done in parallel. The LSTM-AE is frozen for the cold-start stage of training $f_\phi$, but in later stages, we also alternatively finetune the decoder in the LSTM Autoencoder. During Sampling process, $A_0$ and $X_0$, $E_0$, $H_0$, are sampled sequentially. Then the time series embedding $H_0$ is decoded with LSTM-AE. Lastly, $A, X, E, D$ are restored to the input space.

Compared with previous diffusion works that diffuse in raw data space \cite{GBD} or latent space \cite{rombach2022high} only, we achieve a mixture of diffusion spaces via the three-level diffusion model. Also, compared with \cite{rombach2022high} that fixes the autoencoders for diffusion, we further align the embedding space of the autoencoder with the latent space of diffusion by finetuning the decoder along with the training of the diffusion model.

\section{Experiments}\label{sec:exp}
\subsection{Settings}



{\noindent \bf Dataset preparation.} To train and evaluate our proposed diffusion model for generating feasible power grid topologies, we consider two benchmark systems: the IEEE 14-bus transmission grid (U.S.) and the European 36-bus distribution grid representing an urban area \cite{mateo2018european}. For each grid, we generate 1,000 perturbed grid instances using a random-walk-based procedure. Starting from the base grid, we apply degree-preserving edge swaps to change the topology while maintaining node degree distribution. Then, we perform short random walks to extract subgraphs, which are merged back into the perturbed graph to introduce additional local structural variation. Finally, we slightly perturb line impedance values. This process yields diverse grid variants for training. Out of the 1,000 generated instances, 878 variants of the 14-bus system and 813 variants of the 36-bus system are \emph{feasible}: they converge under power flow simulation. The remaining instances are \emph{infeasible}, due to issues such as grid islanding or excessively high branch impedances. Only the feasible variants are used for training the proposed model. Power flow simulations are performed using the PYPOWER library \cite{pypower}, incorporating realistic hourly load profiles from Duquesne Light Company (Pittsburgh). This ensures that the feasibility criterion reflects real-world operational demands, as a feasible topology must be capable of serving realistic load demand patterns. To capture the increasingly dynamic and unpredictable nature of power systems, we also simulate a wide range of structural and load variability across the dataset as case studies to evaluate whether the generated grids can sustain reliable, cost-effective, and resilient performance under diverse operational scenarios.

\begin{figure*}
    \centering
    \vskip -0.1in
    \includegraphics[width=1\linewidth]{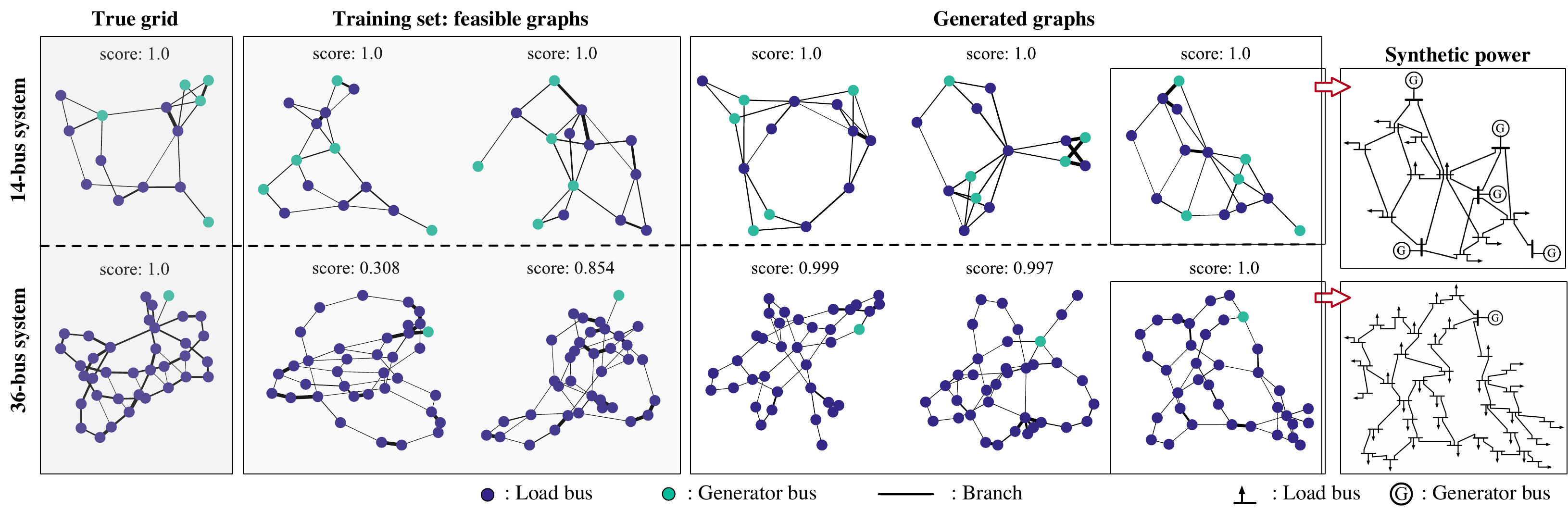}
    \vskip -0.1in
    \caption{Visualization of generated 14-bus and 36-bus power grid samples. Nodes are color-coded by type (generator bus or load bus), and branch impedance is represented via line thickness.}
    \label{fig:comparison}
    \vskip -0.1in
\end{figure*}

{\bf Baselines.} We compared with four state-of-the-art graph diffusion models that can deal with numerical values, including
\begin{itemize}
    \item EDP-GNN \cite{EDP-GNN} designs a permutation invariant, multi-channel graph neural network to model the score function for score matching.
    \item GDSS \cite{GDSS} is a novel continuous-time score-based generative model that model the joint distribution of nodes and edges through a system of stochastic differential equations. 
    \item GruM \cite{GruM} utilizes a mixture of diffusion processes in denoising diffusion for a rapid convergence.
    \item GBD \cite{GBD} applies beta diffusion to graph data to model both continuous and discrete components in graphs.
    \item Grids generated by random-walk perturbation that we used for training and evaluation.
\end{itemize}

{\bf Metrics.} 
This work is motivated by the need for synthetic grid datasets that exhibit both \emph{statistical fidelity} and \emph{operational feasible} given that real-industry grid data is often scarce, privacy-sensitive, and subject to regulatory constraints.
(1) \emph{Fidelity} assesses whether generated topologies and load profiles statistically resemble those in real-world systems. To this end, we report Maximum Mean Discrepancy (MMD) scores across multiple dimensions: the \textbf{Deg.}, \textbf{Clus.}, \textbf{Orbit}, and \textbf{Spec.} metrics correspond to MMDs over degree distributions, clustering coefficients, graphlet orbit counts, and spectral densities, respectively. \textbf{Time.} and \textbf{Attr.} evaluates MMD over time-series bus-level load and edge attributes respectively. Lower MMD values across all metrics indicate closer alignment with the empirical distribution of the training set, and thus stronger realism. 
(2) \emph{Feasibility} captures whether the topology admits a convergent power flow solution with minimal violations. We quantify this using the \textbf{convergence rate} and a \textbf{feasibility score} defined as
\begin{equation}\label{eq:feasibility_score}
    \text{score} = \exp\left(-\tau \cdot \|\mathbf{v}\|_1\right) \in (0,1],
\end{equation}
where $\mathbf{v}$ captures violations of physical and operational constraints, such as voltage magnitude bounds, thermal branch limits, and other system inequalities \cite{pypower}. A complete specification of the constraint terms is provided in Appendix~\ref{app:v:constraint}. $\tau=10^{-5}$ is a scaling factor controlling the sensitivity to violation magnitude; the exponential decay ensures scores near 1 for nearly feasible grids and near 0 for severely infeasible ones.

Additional implementation details and code are provided in Appendix~\ref{sec:app:impl}.

\vspace{-1em}
\subsection{Results and Analysis}

\begin{table*}[t]
    \centering
    \caption{Comparison of comprehensive metrics for graph generations across baseline methods in 36-bus system.}
    \vskip -0.1in
    \begin{tabular}{c|c|c|c|c|c|c|c|c}
        \hline
        Methods & Deg. $\downarrow$ & Clus. $\downarrow$ & Orbit $\downarrow$ & Spec. $\downarrow$ & Time. $\downarrow$ & Attr. (\%) $\downarrow$ & Convergence rate (\%) $\uparrow$ & Feasibility score $\uparrow$ \\
        \hline
        EDP-GNN \cite{EDP-GNN} & 0.4974 & 0.0791 & 0.1944 & 0.3677 & 0.0883 & 0.0833 & 0 & 0 \\
        GruM \cite{GruM} & 0.1842 & 0.1463 & 0.1089 & 0.1783 & 0.0897 & 0.0198 & 0 & 0 \\
        GBD \cite{GBD} & 0.5464 & 0.3053 & 0.2264 & 0.3429 & 0.2167 & 0.0136 & 0 & 0 \\
        GDSS \cite{GDSS} & 0.0000 & 0.0078 & 0.0000 & 0.0085 & 0.0896 & 0.0018 & 67.5 & 0.664 \\
        Random-walk & 0.0000 & 0.0036 & 0.0001 & 0.0033 & 0.0042 & $3\times10^{-5}$ & 81.3 & 0.696 \\
        PowerGrow (Ours) & 0.0006 & 0.0037 & 0.0018 & 0.0061 & 0.0891 & 0.0077 & 98.9 & 0.967 \\
        \hline
    \end{tabular}
    \label{tab:exp:mmd}
\vskip -0.1in
\end{table*}

We begin by visualizing representative grid instances in Figure~\ref{fig:comparison}. The left panel displays samples from the training set of 14-bus and 36-bus system, while the right shows grids generated by our model. Both sets exhibit fully connected topologies without isolated components, satisfying a fundamental requirement for operational power grids. In addition, the generated graphs, similar to those in the training set, exhibit small cycles and meshed substructures that are characteristic of urban distribution networks \cite{xiao2023distribution}, as opposed to trivial tree-like configurations.

\begin{figure}[h]
    \centering
    \vskip -0.1in
    \includegraphics[width=1\linewidth]{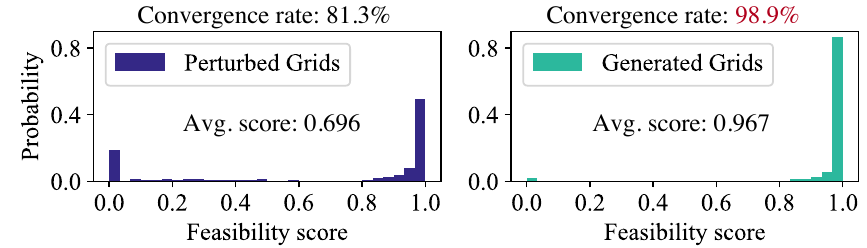}
    \vskip -0.1in
    \caption{Comparison of convergence rate and feasibility scores in the 36-bus system dataset.}
    \label{fig:plots_scores}
    \vskip -0.15in
\end{figure}

To assess operational feasibility, Figure~\ref{fig:plots_scores} compares convergence rates and feasibility scores across two sets: (i) 1,000 random-walk perturbations of the 36-bus system (813 used for training), and (ii) 200 samples from our model. Perturbed variants achieve only an 81.3\% convergence rate and a 0.696 average feasibility score, highlighting the difficulty of preserving physical operability. In contrast, PowerGrow attains a 98.9\% convergence rate and a 0.967 average score, outperforming both perturbed and many training instances. Notably, feasibility is not enforced during training: our diffusion model implicitly learns to generate valid topologies by capturing structural and load patterns from data. These results demonstrate PowerGrow’s ability to synthesize novel yet operationally meaningful grids, addressing a core limitation of prior methods and enabling realistic data generation for downstream tasks such as forecasting, control, and anomaly detection.




Next, we present a comprehensive evaluation in Table~\ref{tab:exp:mmd}, comparing our method against a range of strong baselines. Our model achieves the highest convergence rate and feasibility score, substantially outperforming prior diffusion-based approaches such as GBD \cite{GBD} and GruM \cite{GruM}, both of which fail to generate physically viable topologies. This failure may stem from their reliance on simultaneously learning bus types, edge structures, and attributes without explicitly modeling their interdependencies. In contrast, our hierarchical design enables effective decomposition of this complex generation process, resulting in a better balance between physical validity and statistical realism. Notably, our method achieves consistently low MMD scores across all topological and feature-level metrics, indicating strong alignment with the training distribution. While the random-walk baseline achieves near-zero MMDs by construction, it lacks physical grounding, as reflected in its degraded feasibility metrics. These results underscore the value of incorporating structured generation stages into graph diffusion models for power system synthesis.

\begin{figure}[H]
    \centering
    \includegraphics[width=1\linewidth]{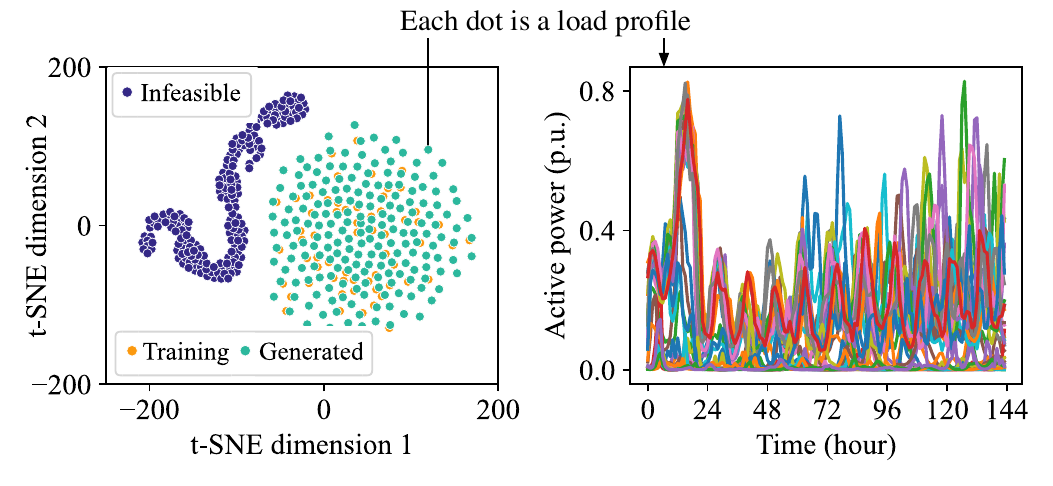}
    \vskip -0.1in
    \caption{{\bf Left}: t-SNE visualization of generated load profile latents in the 36-bus system. {\bf Right}: bus-level time-series active power injections.}
    \label{fig:t-SNE-plot}
    \vskip -0.1in
\end{figure}

We then evaluate the quality of the generated time-series load profiles. The left panel of Figure~\ref{fig:t-SNE-plot} visualizes the active and reactive power time-series generated by our model for each load bus. The active power represents real electricity consumption, while the reactive power is essential for voltage regulation and system stability. As shown, the generated profiles exhibit realistic temporal variability and oscillatory behavior, suggesting that they are sufficiently rich to support downstream tasks such as forecasting, control, and stability analysis. 

To further assess the realism of these profiles, we apply Principal Component Analysis (PCA) for dimensionality reduction, followed by t-SNE to embed the high-dimensional power flow trajectories into two dimensions for visualization. We compare the resulting embeddings across three sets: time-series profiles from the training set, from our generated grids, and from infeasible grids produced via random-walk perturbations. The right panel of Figure~\ref{fig:t-SNE-plot} shows that the generated profiles cluster tightly with those of the feasible training set, indicating strong alignment in both structural and dynamic characteristics. In contrast, the infeasible grids form a dispersed and distinct cluster, highlighting the instability and inconsistency of their simulated power behavior. It demonstrates that our model not only generates topologically valid grids but also supports realistic, high-fidelity load dynamics, which are essential for practical deployment.

We also compare the sampling efficiency of PowerGrow against baselines, each tasked with generating a complete power grid sample including both the topology and corresponding load profile. Table~\ref{tab:sampling_speed} reports the average sampling time per grid, measured over 200 runs on the 36-bus system. Among the baselines, PowerGrow achieves efficient end-to-end generation of both physically feasible topologies and dynamic load profiles in 0.525s per sample. This represents up to a 31\% speed improvement over baseline diffusion models such as EDP-GNN and GruM. Compared to the optimization-based method \cite{yan2022active}, which solves a constrained power flow problem to derive feasible loads, PowerGrow is over 85× faster.

\begin{table}[h]
\centering
\caption{Average sampling time per grid (topology + load).}
\label{tab:sampling_speed}
\vskip -0.1in
\begin{tabular}{l|c}
\hline
\textbf{Method} & \textbf{Time per sample (sec)} \\
\hline
PowerGrow (Ours) & 0.525 \\
EDP-GNN \cite{EDP-GNN} & 0.635 \\
GruM \cite{GruM} & 0.765 \\
GDSS \cite{GDSS} & 2.620 \\
GBD \cite{GBD} & 0.501\\
OPT-based \cite{yan2022active} & 46.255 \\
\hline
\end{tabular}
\vskip -0.1in
\end{table}




\subsection{Ablation Study}

\begin{figure}[h]
    \centering
    \vskip -0.1in
    \includegraphics[width=1\linewidth]{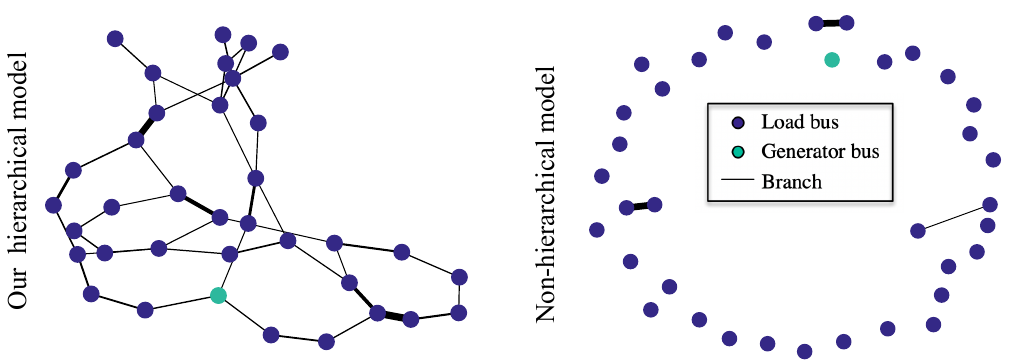}
    \vskip -0.05in
    \caption{Generated topologies of our hierarchical diffusion model and an ablated version that applies non-hierarchical diffusion model in the 36-bus system.}
    \label{fig:ablation}
    \vskip -0.1in
\end{figure}

To assess the effectiveness of the proposed hierarchical generation strategy in PowerGrow, Figure~\ref{fig:ablation} compares the generated topologies from our full model and a variant that removes the hierarchical generation structure. In the variant, all components (node types, topology, and edge attributes) are generated jointly in a single diffusion process without enforcing causal ordering. Despite being trained for the same number of epochs (60,000), this baseline produces a fragmented network with only 3 branches across 36 buses, lacking basic connectivity. In contrast, our hierarchical approach yields a well-structured and fully connected grid. This result reinforces our design choice: by decomposing the generation process into causally structured stages, our model aligns with the physical generation logic of power systems and avoids complex distribution learn. In contrast, the joint-generation variant lacks this guidance, making it prone to disconnected topologies due to the compounded difficulty of learning multiple interdependent factors simultaneously.

\vspace{-1em}
\subsection{Trajectory of Generated Grid Topology}
\label{sec:app:tra-topology}

To better understand the convergence behavior of our proposed diffusion model, we visualize the generated topologies at different stages of the training process. Specifically, we plot the graph topology generated by the model every 1200 training epochs Figure~\ref{fig:tra-topology}. Load profile generation trajectory is shown in Appendix \ref{sec:app:tra-load}.

The early-stage generation (e.g., epoch = 1200) often results in disconnected graphs with multiple isolated buses or ``islands''. Such disconnected structures are undesirable for power grid planning, as they indicate infeasible networks. As training progresses, the model learns to generate more coherent and physically plausible topologies. By later epochs (e.g., epoch = 9600 and beyond), the generated graphs exhibit fully connected and structurally coherent layouts, consistently achieving feasibility scores of 1. This indicates convergence toward operationally valid grid topologies. Notably, this evolution suggests that the diffusion model progressively internalizes structural and physical constraints, despite the absence of any explicit connectivity enforcement during generation.

\begin{figure}[h]
    \centering
    \vskip -0.1in
    \includegraphics[width=0.9\linewidth]{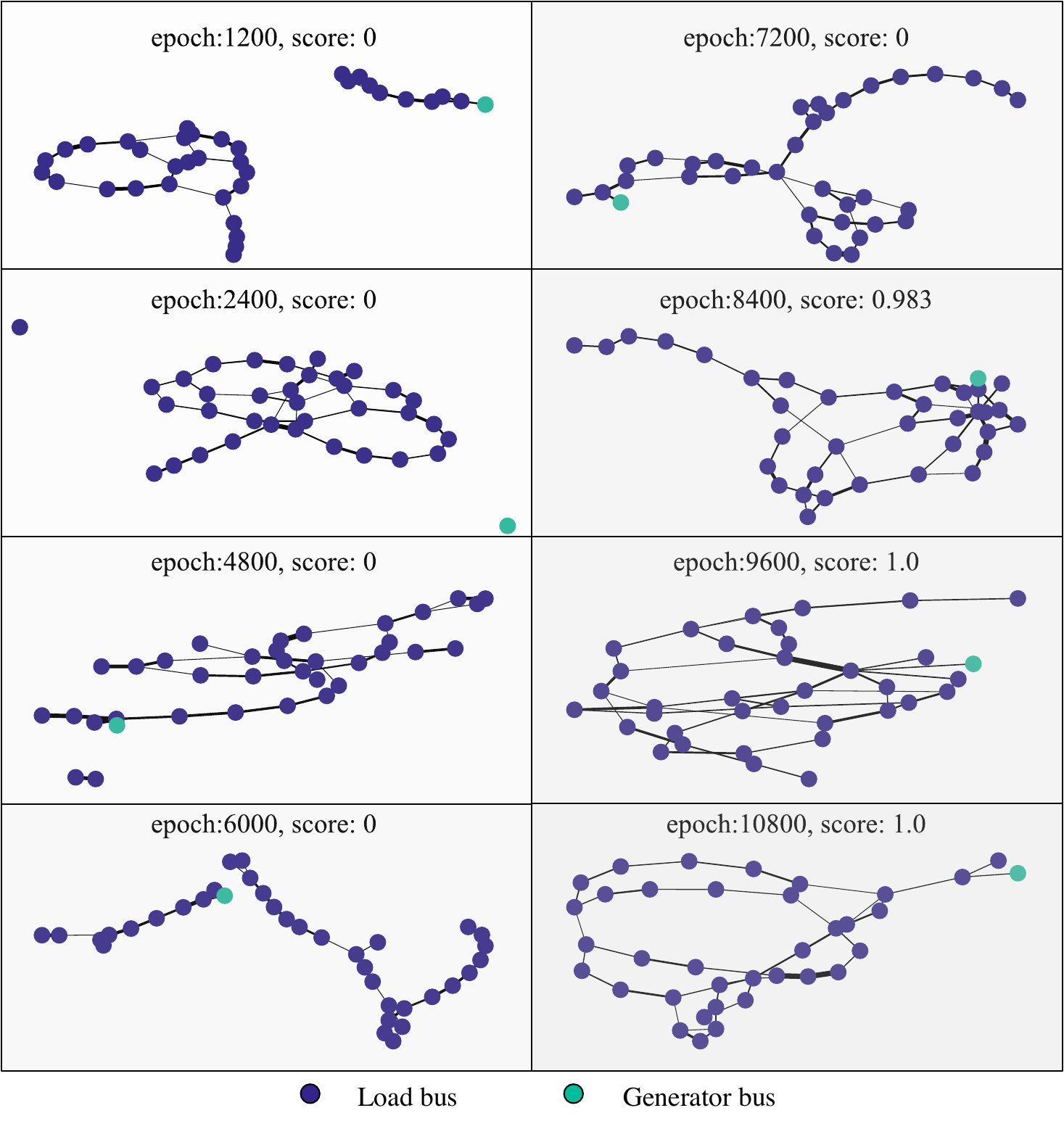}
    \vskip -0.1in
    \caption{Evolution of generated topology across training epochs in the 36-bus dataset.}
    \vskip -0.1in
    \label{fig:tra-topology}
\end{figure}

\subsection{Case Study: Economic Efficiency}

\begin{figure}[h]
    \centering
    \includegraphics[width=1\linewidth]{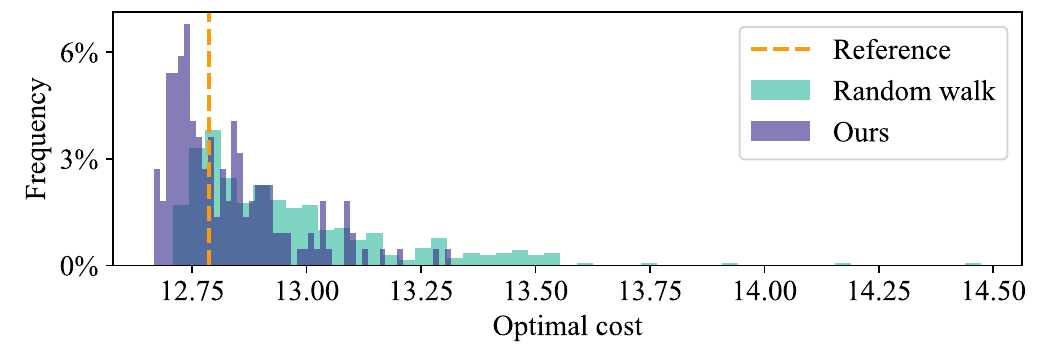}
    \vskip -0.15in
    \caption{Histogram of ACOPF costs for grids generated by our diffusion model and random-walk procedure. The vertical dashed line is the cost of the true 36-bus system.}
    \label{fig:OPF-cost}
    \vskip -0.1in
\end{figure}

We assess the real-world value of the generated power grid topologies using a case study based on AC Optimal Power Flow (ACOPF) analysis. Each graph is assessed under a fixed load demand and generator setup, with the resulting objective cost (reflecting total generation expense) serving as a proxy for economic efficiency. We compare ACOPF cost distributions of topologies produced by our diffusion model to those from a random walk baseline, which perturbs a reference grid via stochastic edge modifications. As illustrated in Figure \ref{fig:OPF-cost}, diffusion-generated topologies consistently yield lower costs. Notably, many of the topologies discovered by the diffusion model also outperform the reference true grid. This demonstrates the model’s capacity to traverse topological regions and discover more efficient power dispatches. Such capability is especially valuable for grid planning and expansion, where minimizing long-term operational cost is paramount.

\subsection{Case Study: N-1 Contingency Resilience}

Another critical characteristic of a plausible power grid is its ability to maintain operational stability under component failures. To evaluate this property, we assess the resilience of each generated topology under the N-1 contingency criterion, which requires the system to remain feasible following the failure of any single transmission line. For each graph, we simulate the power outage of every line by disabling it and solving the resulting AC power flow. We evaluate the \emph{N-1 resilience rate}: the proportion of single-line contingencies under which power flow converges successfully. As shown in Table~\ref{tab:n1_resilience}, our diffusion-based model achieves a resilience rate of 73.19\%, surpassing both the true 36-bus system (72.91\%) and random walk baselines (65.24\%). This demonstrates PowerGrow’s ability to generate topologies that are structurally robust, as an essential property for reliable grid planning.


\begin{table}[h]
\centering
\caption{N-1 resilience rate (\%) across different methods on the 36-bus dataset.}
\vskip -0.1in
\begin{tabular}{l|ccc}
\hline
Methods & Reference Grid & Random Walk & Ours \\
\hline
Rate (\%) & 72.91 & 65.24 & 73.19 \\
\hline
\end{tabular}
\label{tab:n1_resilience}
\vskip -0.15in
\end{table}

\subsection{Case Study: Load Shedding Under Stress}
\begin{figure}[h]
    \centering
    \vskip -0.1in
    \includegraphics[width=1\linewidth]{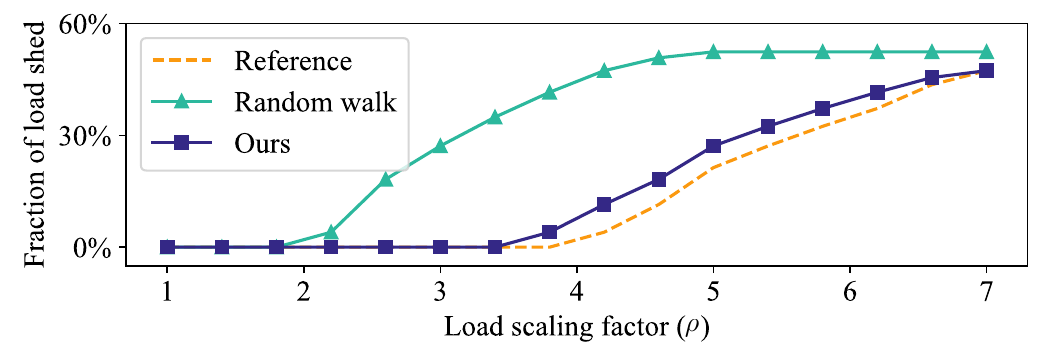}
    \vskip -0.1in
    \caption{Fraction of load shed against load scaling factor $\rho$ in the 14-bus dataset.}
    \label{fig:load_shed}
    \vskip -0.1in
\end{figure}

We further evaluate the stress resilience of the generated topologies, motivated by real-world scenarios such as high-demand conditions or partial generation loss. Specifically, we scale the original load demand of 14-bus system by a factor $\rho>1$ to measure the minimum fraction of total demand that must be shed to achieve a feasible AC power flow. For each topology, we iteratively reduce load at high-demand buses until convergence is restored, using the resulting shed fraction as a measure of the system’s ability to maintain service under stress. Figure~\ref{fig:load_shed} shows the fraction of load shed as a function of $\rho$ for the reference grid, the random walk-generated grids (averaged), and the diffusion-generated grids (averaged). The random walk method begins requiring load shedding before $\rho=2$, indicating limited structural resilience. In contrast, our diffusion-generated grids exhibit significantly improved behavior, closely tracking the reference grid up to $\rho\approx3.5$, after which both methods begin to shed load gradually. It indicates that the proposed model generates topologies that are not only feasible and economically efficient but also structurally robust under stress.

\section{Conclusion and Future Work}
\label{sec:conclusion}
In this work, we introduce PowerGrow, a novel co-generation framework for power grid generation and power loads synthesis. To the best of our knowledge, we are the first to leverage their inherent interdependency for the joint co-generation of power grid structure and load profiles with neural networks. For efficient time series generation, we propose a three-stage training paradigm that trains a LSTM autoencoder to project time series data into latent space and aligns the latent space with diffusion output space via finetuning. To bypass the joint generation task of heterogeneous features in power grids, we propose a three-level hierarchical graph diffusion framework for diffusion in the mixture of data and latent space. Our experiments demonstrate the high fidelity and feasibility of our generated power grids. In the future, we will automate feasibility evaluation and encourage efficient explorations using reinforcement learning or large language models.

\clearpage


\bibliographystyle{ACM-Reference-Format}
\bibliography{main}

\appendix
\section{Construction of the Violation Vector}
\label{app:v:constraint}

To quantify the physical plausibility of generated power grids, we define a continuous feasibility score based on the magnitude of operational constraint violations. Let the result of an AC power flow simulation be represented by voltage magnitudes $V_i$, phase angles $\theta_i$, and complex power flows $S_{ij} = P_{ij} + j Q_{ij}$ on branch $(i, j)$. The constraint violation vector $\mathbf{v}$ aggregates violation magnitudes for key physical constraints:
\begin{itemize}
\item Voltage magnitude limits. For each bus $i$, voltage must remain within its permissible bounds: $V_i^{\min} \leq V_i \leq V_i^{\max}$, which indicates the constraint violation as 
\begin{align}
v_i^{(V)} = \max\Big(0, V_i - V_i^{\max} \Big) + \max\left(0, V_i^{\min} - V_i\right).
\end{align}

\item Branch flow limits. Each transmission line must respect its thermal constraint on apparent power: $|S_{ij}| \leq S_{ij}^{\max}$, which indicates the constraint violation as
\begin{align}
v_{ij}^{(S)} = \max\left(0, |S_{ij}| - S_{ij}^{\max}\right).
\end{align}

\item Generator capacity limits. When available, generator outputs must lie within capacity constraints: $P_i^{\min} \leq P_i^{\text{gen}} \leq P_i^{\max}$ and $Q_i^{\min} \leq Q_i^{\text{gen}} \leq Q_i^{\max}$, which indicates the constraint violation as
\begin{align}
v_i^{(P)} &= \max\left(0, P_i^{\text{gen}} - P_i^{\max}\right) + \max\left(0, P_i^{\min} - P_i^{\text{gen}}\right),\\
v_i^{(Q)} &= \max\left(0, Q_i^{\text{gen}} - Q_i^{\max}\right) + \max\left(0, Q_i^{\min} - Q_i^{\text{gen}}\right).
\end{align}
\end{itemize}

The complete violation vector is given by:
\[
\mathbf{v} = \left[ \{v_i^{(V)}\}_i, \{v_{ij}^{(S)}\}_{(i,j)}, \{v_i^{(P)}\}_i, \{v_i^{(Q)}\}_i \right],
\]
where generator terms $v^{(P)}$ and $v^{(Q)}$ are included only if generator limits are available in the dataset. In experiments, the vector $\mathbf{v}$ is constructed using the raw constraint violation array from the \texttt{runopf} routine in PYPOWER~\cite{pypower}, accessed via \texttt{results["raw"]["g"]}. This formulation enables a differentiable and fine-grained feasibility score, facilitating smooth evaluation of physical realism in generated grids.

\begin{algorithm}
\caption{Training pipeline of PowerGrow}
\begin{algorithmic}[1]
\REQUIRE
    \STATEx Number of diffusion steps $T$, concentration parameters $\eta_X$ and $\eta_A$, graph transformer models $f_\theta(), f_\gamma(), f_\phi()$, LSTM Autoencoder $g()$, input graphs in batches $\mathbb{B}$, number of epochs $E$, normalization parameters $\mathbf{w}=(w_A\ w_X\ w_E\ w_D), \mathbf{b}=(w_A\ w_X\ w_E\ w_D)$, number of cold-start epochs $M$
    \FOR{epoch from $1$ to $E$}
        \FOR{$(A, X, E, D)\in\mathbb{B}$}
            \STATE Normalize $D_0\leftarrow w_DD+b_D$
            \STATE Reconstruct $\hat{D}_0\leftarrow g(D_0)$
            \STATE Update $g()$ with $\mathcal{L} = \text{MSE}(D_0, \hat D_0)$
        \ENDFOR
    \ENDFOR
    \FOR{epoch from $1$ to $E$}
        \FOR{$(A, X, E, D)\in\mathbb{B}$}
            \STATE Normalize $(A_0\ X_0\ E_0\ D_0)\leftarrow \mathbf{w}(A\ X\ E\ D)+\mathbf b$
            \STATE $t\sim\text{Unif}(1,...,T)$
            \STATE Sample $A_t, X_t$ with Eq.~\eqref{eq:gbd:forward}
            \STATE Predict $\hat{A}_0, \hat{X}_0\leftarrow f_\theta(A_t, X_t, t)$
            \STATE Update $f_\theta()$ with KLUB loss Eq.~\eqref{eq:klub} on $A$ and $X$
        \ENDFOR
    \ENDFOR
    \FOR{epoch from $1$ to $E$}
        \FOR{$(A, X, E, D)\in\mathbb{B}$}
            \STATE Normalize $(A_0\ X_0\ E_0\ D_0)\leftarrow \mathbf{w}(A\ X\ E\ D)+\mathbf b$
            \STATE $t\sim\text{Unif}(1,...,T)$
            \STATE Sample $E_t$ with Eq.~\eqref{eq:gbd:forward}
            \STATE Predict $\hat{E}_0\leftarrow f_\gamma(A_0, X_0, E_t, t)$
            \STATE Update $f_\gamma()$ with KLUB loss Eq.~\eqref{eq:klub} on $E$
        \ENDFOR
    \ENDFOR
    \FOR{epoch from $1$ to $E$}
        \FOR{$(A, X, E, D)\in\mathbb{B}$}
            \STATE Normalize $(A_0\ X_0\ E_0\ D_0)\leftarrow \mathbf{w}(A\ X\ E\ D)+\mathbf b$
            \STATE $t\sim\text{Unif}(1,...,T)$
            \STATE $H_0\leftarrow g_{enc}(D_0)$
            \STATE Sample $H_t$ with Eq.~\eqref{eq:gbd:forward}
            \STATE Predict $\hat{H}_0\leftarrow f_\phi(A_0, X_0, E_0, H_t, t)$
            \STATE Update $f_\phi()$ with KLUB loss Eq.~\eqref{eq:klub} on $H$
            \IF{epoch $> M$}
                \STATE $\hat{D}_0\leftarrow g_{dec}(\hat H_0)$
                \STATE Update $g()$ with $\text{MSE}(D_0, \hat D_0)$
            \ENDIF
        \ENDFOR
    \ENDFOR
\end{algorithmic}
\label{alg:train}
\end{algorithm}

\begin{algorithm}
\caption{Sampling Process of PowerGrow}
\begin{algorithmic}[1]
\REQUIRE
    \STATEx Number of diffusion steps $T$, concentration parameters $\eta_X$ and $\eta_A$, graph transformer models $f_\theta(), f_\gamma(), f_\phi()$, LSTM Autoencoder $g()$, normalization parameters $\mathbf{w}=(w_A\ w_X\ w_E\ w_D), \mathbf{b}=(w_A\ w_X\ w_E\ w_D)$
    \STATE Sample $A_T\sim\mathtt{Beta}(\eta_A\alpha_T\mathbb{E}[A_0])$, $E_T\sim\mathtt{Beta}(\eta_A\alpha_T\mathbb{E}[E_0])$
    \STATE Sample $X_T\sim\mathtt{Beta}(\eta_X\alpha_T\mathbb{E}[X_0])$, $H_T\sim\mathtt{Beta}(\eta_X\alpha_T\mathbb{E}[H_0])$
    \FOR{t from $T$ to $1$}
        \STATE $\hat A'_0, \hat X'_0\leftarrow f_\theta(A_t, X_t, t)$
        \STATE $\hat A_0\leftarrow w_A\hat A'_0+b_A, \hat X_0\leftarrow w_X\hat X'_0+b_X$
        \STATE Sample $A_{t-1}, X_{t-1}$ via Eq.~\eqref{eq:sample:add}
    \ENDFOR
    \FOR{t from $T$ to $1$}
        \STATE $\hat E'_0\leftarrow f_\gamma(A_0, X_0, E_t, t)$
        \STATE $\hat E_0\leftarrow w_E\hat E'_0+b_E$
        \STATE Sample $E_{t-1}$ via Eq.~\eqref{eq:sample:add}
    \ENDFOR
    \FOR{t from $T$ to $1$}
        \STATE $\hat H'_0\leftarrow f_\phi(A_0, X_0, E_0, H_t, t)$
        \STATE Sample $H_{t-1}$ via Eq.~\eqref{eq:sample:add}
    \ENDFOR
    \STATE $D_0\leftarrow g_{dec}(H_0)$
    \STATE Output $(A\ X\ E\ D) = \big((A_0\ X_0\ E_0\ D_0) - \mathbf{b}\big)/\mathbf{w}$
\end{algorithmic}
\label{alg:sample}
\end{algorithm}

\section{Implementation Details}\label{sec:app:impl}
The overall training and sampling pipeline are summarized in Alg.~\ref{alg:train} and Alg.~\ref{alg:sample}. Our codes and dataset are available at \href{https://github.com/xinyuu-he/PowerGrow}{Github}.

\textbf{Data Preprocessing.} We normalize the branch attributes to bounded range $[0,1]$, for example, for branch resistances in the raw dataset $\{\tilde{r}|\tilde{r}\in\mathbb{R}^{M}\}$, where $M$ is the number of branches in each graph, it is normalized by
\begin{equation}
    r = (\tilde{r}-\mu_r)*\sigma_r + 0.5
\end{equation}
where $\mu_r, \sigma_r$ are predefined parameters so that $\mu_r=mean_{\tilde{r}}\big(mean(\tilde{r})\big)$, and $max(r)$ is close or equal to $1$. Time-series load profile are also rescaled, e.g., for real-part of load profiles $\{\tilde{P}|\tilde{P}\in\mathbb{R}^{N\times T}\}$, we apply $P = \tilde{P}\cdot \sigma_P$ so that $max(P)$ is close to $1$.

\textbf{Logit domain computation.} To ensure numerical accuracy, we follow \cite{GBD, zhou2023beta} to train in the logit space. For example, in logit domain, Eq.~\eqref{eq:sample:add} is expressed as
\begin{equation}
    \text{logit}(A_{k-1}) = \ln\big(e^{\text{logit}(A_k)}+e^{\text{logit}(P_{A,k})}+e^{\text{logit}(A_k)+\text{logit}(P_{A,k})}\big).
\end{equation}

\textbf{Baselines.} We run their official implementations for baselines. GruM, EDP-GNN, GDSS only generate one-dimensional edge feature (adjacency matrix) in their original version, therefore we extend the graph transformers in their codes to accept multi-dimensional features by considering the dimensions as different channels and copying the operations on each dimension. EDP-GNN does not generate node attributes as well, therefore we randomly generate load profiles through our pretrained decoder in LSTM-AE with random noise as inputs. As EDP-GNN reaches the lowest Time. MMD in all baselines, the effectiveness of our LSTM-AE for time-series generation is further confirmed.

\newpage
\section{Evolution of Load Profile Generation During Diffusion Training}
\label{sec:app:tra-load}
In addition to topology evolution, we also investigate how the generated time-series load profiles improve over the training of diffusion training. We focus on the real power consumption vector $P_{\text{vec}}$, expressed in per unit (p.u.), and visualize the generated time-series curves every 1200 epochs.

\begin{figure}[H]
    \centering
    \includegraphics[width=1\linewidth]{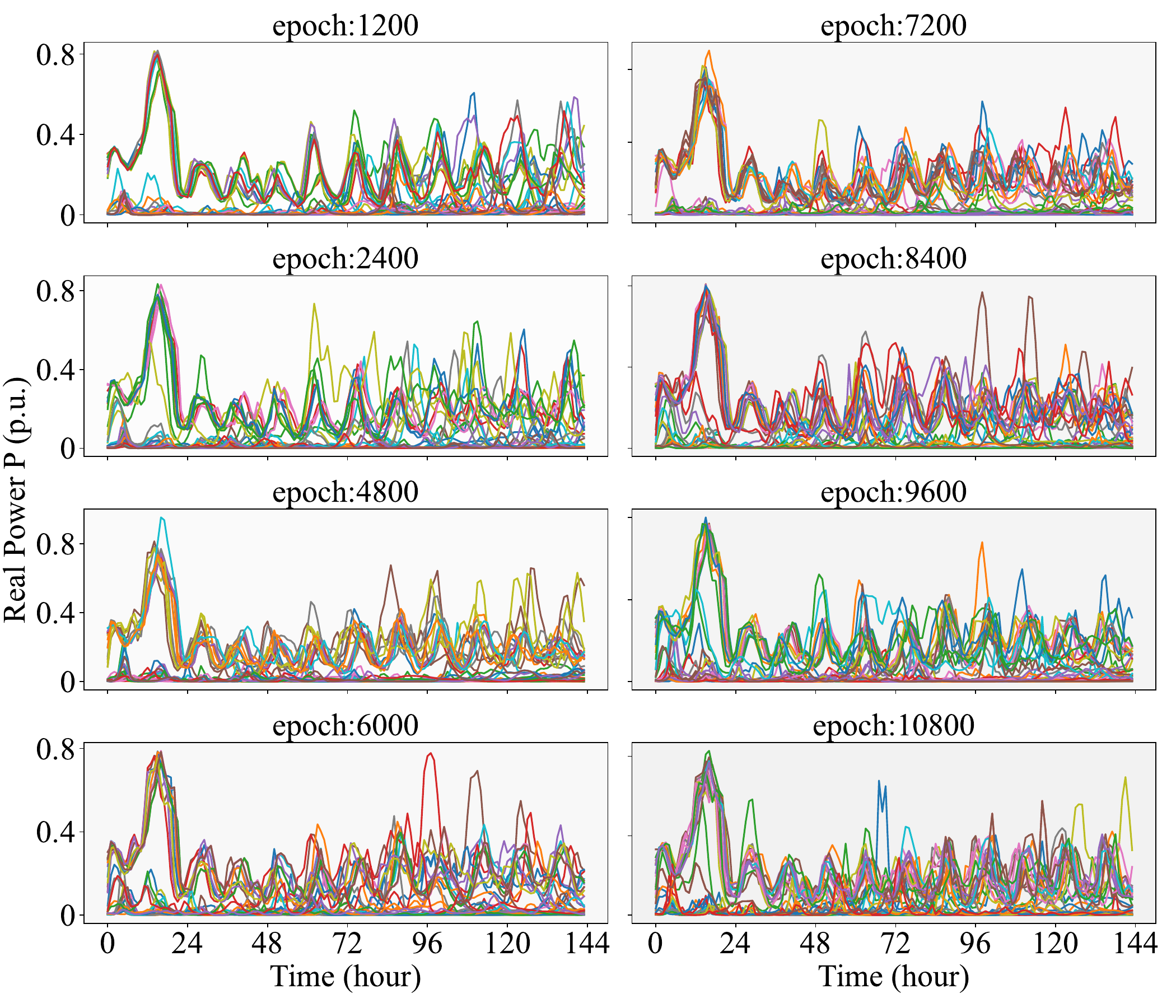}
    \caption{Evolution of generated real power (p.u.) time-series across training epochs in the 36-bus dataset.}
    \label{fig:tra-timeseries}
\end{figure}

Figure~\ref{fig:tra-timeseries} presents the generated $P_{\text{vec}}$ sequences across training stages. Early in training, the generated profiles exhibit limited variability and lack meaningful temporal patterns. However, it already exhibits clear periodical patterns which confirms effectiveness of pretrained LAST-AE. As training advances, the model begins to produce more structured and realistic time-series patterns, capturing typical fluctuations found in real-world load behavior. By the final stages of training, the time-series exhibit smooth yet diverse trajectories that align with expected residential or commercial load dynamics. This progression highlights the ability of the diffusion model to learn both spatial and temporal structures jointly.

\end{document}